\documentclass[11pt]{article}
\usepackage{hyphenat}
\usepackage[margin=1in]{geometry}
\usepackage{amsmath,amssymb}
\usepackage{booktabs}
\usepackage{microtype}
\usepackage{enumitem}
\usepackage[T1]{fontenc}
\usepackage{inconsolata}
\usepackage{pgfplots}
\pgfplotsset{compat=1.18}
\usepackage{float}
\usepackage{hyperref}

\title{\vspace{-0.5em}Semantic Fusion with Fuzzy-Membership Features for Controllable Language Modelling}
\author{Yongchao Huang\thanks{Corresponding author: \texttt{yongchao.huang@abdn.ac.uk}}, Hassan Raza}
\date{12 Sept. 2025}

\begin{document}
\maketitle

\begin{abstract}
We propose \textit{semantic fusion}, a lightweight scheme that augments a Transformer language model (LM) with a parallel, fuzzy-membership feature channel that encodes token-level semantics. Each token is represented by a vector of interpretable features (e.g. part-of-speech cues, shallow roles, boundary flags, sentiment polarity and strength) whose values are graded degrees from differentiable membership functions (e.g. power kernels). These per-token vectors form a sentence-level semantic matrix fused via a gated adapter into the LM. Training uses standard next-token prediction, an auxiliary loss that reconstructs the semantic features from hidden states, and a lightweight “uniformizer” that regularizes adjective-class distributions. On a synthetic two-clause corpus with held-out adjectives for out-of-distribution (OOD) control, semantic fusion improves perplexity and enables precise, user-controllable generation of polarity and punctuation while maintaining model simplicity. This approach adds only small overhead, remains fully compatible with tied input-output embeddings, and provides an interpretable pathway for conditioned natural language generation.
\end{abstract}

\vspace{-0.5em}
\section{Background}

Conventional token embeddings entangle many factors (syntax, semantics, style), which can limit both controllability and interpretability \cite{mikolov2013efficient,peters2018deep,ethayarajh2019contextual}. We ask whether \textit{interpretable} semantics\footnote{We use “semantic” to denote interpretable token-level features (e.g. roles, POS-ish cues, boundary/style, and sentiment polarity/strength). This differs from formal, truth-conditional semantics in logic/knowledge representation.} can be exposed as an auxiliary channel that both improves modeling accuracy and provides handles for \textit{controllable} generation \cite{keskar2019ctrl,dathathri2020plug,krause2021gedi,hokamp2017lexically}. In standard neural LMs, each token $w\in\mathcal{V}$ is represented by an index and mapped to a dense vector $e=E[w]$ via an embedding table $E\in\mathbb{R}^{|\mathcal{V}|\times d}$; these embeddings, combined with positional signals (e.g. sinusoidal \cite{vaswani2017attention} or rotary \cite{su2023roformer}), are processed by a Transformer \cite{vaswani2017attention}. While effective, this representation is \textit{implicit}: part-of-speech, syntactic role, polarity, and stylistic cues are folded into $e$ and must be discovered from data \cite{tenney2019bert,hewitt2019structural}, and conditioning is typically imposed externally through prompts or control codes \cite{keskar2019ctrl}, whose effects can be brittle.

Fuzzy logic offers a complementary view in which concepts are expressed by \textit{membership functions} that assign graded degrees of truth \cite{zadeh1965fuzzy}. A fuzzy set on a universe $\mathcal{U}$ is defined by $\mu:\mathcal{U} \to [0,1]$, and linguistic categories such as \textit{low/medium/high intensity} are modeled by overlapping memberships (e.g. triangular, trapezoidal, or Gaussian \cite{klir1995fuzzy}) that softly partition a scalar variable. We leverage this idea at the token level: predicates such as \texttt{is\_adj}, \texttt{pos\_high}, or \texttt{str\_med} are encoded as fuzzy memberships and used as differentiable targets. Concretely, for a scalar signal $x$ (e.g. sentiment magnitude) and center $c$ with temperature $\tau$, we use the membership function
\[
\mu(x;c,\tau) = 0.9^{\,|x-c|/\tau}\in(0,1]
\]
which decreases smoothly with distance and is amenable to gradient-based learning.

We then fuse these graded, interpretable features with the text stream, so the model learns next-token prediction while aligning hidden states to an explicit semantic scaffold. This brings three benefits: (i) a low-variance supervisory signal that guides the backbone toward useful abstractions at the positions where they matter (e.g. adjective and punctuation slots), (ii) \textit{continuous control variables} (real-valued scalars in $[0,1]$) for test-time steering (e.g. increasing \texttt{str\_high} smoothly raises the probability of higher-intensity completions), and (iii) a bias towards \textit{class-level} generalization (e.g. the set of positive adjectives) rather than memorization of individual words. Importantly, the predicates used for training are the same ones used for control at inference, aligning supervision with steering and improving both perplexity and the reliability of conditioned generation.

\vspace{-0.25em}
\section{Method} \label{sec:method}

\paragraph{Fuzzy semantic features.}
Let a sentence have length $L$ and tokens $\{w_t\}_{t=1}^L$.
For each position $t$ we build a semantic feature vector $s_t\in[0,1]^F$, where $F$ is the number of predicates in a fixed \textit{feature bank}:
POS-like cues (\texttt{is\_noun}, \texttt{is\_verb}, \texttt{is\_adj}),
shallow syntax (\texttt{is\_subject}, \texttt{is\_object}, \texttt{is\_head}),
boundary flags (\texttt{is\_bos}, \texttt{is\_eos}, \texttt{is\_comma}, \texttt{is\_question}),
sentiment triplets (\texttt{pos\_low/med/high}, \texttt{neg\_low/med/high}),
a strength triplet (\texttt{str\_low/med/high}),
and light discourse/style (\texttt{coref\_subject}, \texttt{is\_capitalized}, \texttt{is\_pronoun}).
Binary predicates take values in $\{0,1\}$; graded predicates use a \textit{power-law triangular} membership function.
Given any scalar attribute value $x\in[0,1]$, a bandwidth $\tau > 0$, and centers $\mathcal{C}=\{c_1,c_2,c_3\}$, we define
\begin{equation}
\mu(x;c,\tau) = 0.9^{ |x-c|/\tau}\in(0,1],\qquad
\mathrm{Tri}(x;\mathcal{C},\tau) = \big[\mu(x;c_1,\tau), \mu(x;c_2,\tau), \mu(x;c_3,\tau)\big]
\end{equation}
We fix $\mathcal{C}=\{0.2,0.6,1.0\}$ and $\tau=0.35$ so that \texttt{low/med/high} correspond to the three centers.

\textit{Sentiment instantiation.}
Let $s^\star(w_t) \in \{-1,0,+1\}$ be a token-level polarity score derived from the adjective lexicon (positive/negative/neutral).
We decompose it into nonnegative channels
$x_{\mathrm{pos}}(t)=\max\{0, s^\star(w_t)\}$ and $x_{\mathrm{neg}}(t)=\max\{0, -s^\star(w_t)\}$,
and set
\[
(\texttt{pos\_low},\texttt{pos\_med},\texttt{pos\_high})_t=\mathrm{Tri} \big(x_{\mathrm{pos}}(t);\mathcal{C},\tau\big),\qquad
(\texttt{neg\_low},\texttt{neg\_med},\texttt{neg\_high})_t=\mathrm{Tri} \big(x_{\mathrm{neg}}(t);\mathcal{C},\tau\big)
\]

\textit{Strength instantiation.}
Let $r^\star(w_t)\in[0,1]$ encode intensity (e.g.\ \texttt{slightly}$\mapsto0.2$, \texttt{moderately}$\mapsto0.5$, \texttt{very}$\mapsto0.8$, \texttt{extremely}$\mapsto1.0$), with an optional $+0.2$ nudge (capped at $1$) when the final punctuation is “!”. 
At the adjective slot, $r^\star$ is derived from the immediately preceding intensifier.
We set
\[
(\texttt{str\_low},\texttt{str\_med},\texttt{str\_high})_t=\mathrm{Tri}\big(r^\star(w_t);\mathcal{C},\tau\big)
\]

Stacking the per-token vectors yields the sentence-level \textit{semantic matrix}
$S=\begin{bmatrix}s_1^\top;\dots;s_L^\top\end{bmatrix}\in[0,1]^{L\times F}$,
which we fuse with the LM.

\paragraph{Gated semantic fusion.}
Let $x_{1:L}$ be token ids over a vocabulary $\mathcal{V}$ of size $V=\lvert\mathcal{V}\rvert$ (including specials \texttt{<bos>}, \texttt{<eos>}, \texttt{<pad>}).
Let $E \in \mathbb{R}^{V\times d}$ be the embedding table, $e_t=E[x_t]\in\mathbb{R}^d$ the standard token embedding, and $s_t \in [0,1]^F$ the fuzzy-membership feature vector at position $t$ (with $F$ the number of predicates in the feature bank).
We project semantics with $u_t=W_s s_t\in\mathbb{R}^d$, compute a token-conditioned gate, and fuse additively:
\begin{align}
g_t &= \sigma \big(W_g [e_t; s_t]\big)\in(0,1)^d, \\
h_t^{(0)} &= e_t + u_t + g_t \odot u_t, \\
H &= \mathrm{TransformerEnc} \left(\mathrm{PosEnc}(h_{1:L}^{(0)})\right)
\end{align}
where $[ \cdot;\cdot ]$ denotes concatenation and $\odot$ the elementwise (Hadamard) product.
The term $g_t \odot u_t$ acts as a per-dimension gate on the semantic projection: since $h_t^{(0)} = e_t + (1+g_t)\odot u_t$ with $g_t \in (0,1)^d$, the semantic contribution is \textit{amplified}\footnote{For example, if $u_{t,j}=0.30$ and $g_{t,j}=0.8$, then the semantic contribution on dimension $j$ becomes $0.30 + 0.8\times 0.30 = 0.54$; if $g_{t,j} \approx 0$, it stays near $0.30$. Because $g_t=\sigma(W_g[e_t;s_t])$, the gate is token- and feature-conditioned and tends to be higher at \textsc{INTENS}/\textsc{ADJ}/\textsc{PUNCT} slots.} by a factor in $(1,2)$ in each dimension (approaching $1$ when $g_t \approx 0$ and $2$ when $g_t \approx 1$).
The LM head is weight-tied\footnote{By “tied input-output embeddings” we reuse the input embedding matrix $E\in\mathbb{R}^{V\times d}$ as the output softmax weight, i.e., $\mathrm{logits}_t = H_t E^\top + b$. Our fusion keeps the hidden size $d$ unchanged and injects the semantic projection $W_s s_t\in\mathbb{R}^d$ \textit{before} the Transformer, so the tied LM head ($E^\top$) remains valid. This preserves the parameter savings and regularization benefits of tying while adding the semantic channel, and no extra vocabulary-sized output layer is introduced.}: $p(y_t\mid x_{\le t})=\mathrm{softmax}(E^\top H_t)$.
An auxiliary head predicts semantics $\hat{s}_t=\sigma(\mathrm{MLP}(H_t))$, aligning hidden states with interpretable features.

\textit{Fusion} here refers to early, gated integration of two parallel token representations (i.e. the usual, learned text embedding $e_t$ and the explicit fuzzy-semantic feature vector $s_t$) into a single representation $h_t^{(0)}$ consumed by the Transformer.
The gate $g_t$ learns \textit{how much} semantic signal to inject per dimension and position (e.g. near \textsc{INTENS}/\textsc{ADJ}/\textsc{PUNCT} slots), while the residual path $e_t{+}u_t$ preserves a strong fallback when semantics are unhelpful.
This design yields both better perplexity (the model gets low-variance cues when they matter) and controllability (the same predicates used at training time are available to steer decoding at test time).

\paragraph{Training objective: label smoothing + auxiliary + uniformizer losses.}
Let $y_{1:L}$ be the target next-token sequence, and let $p_\theta(\cdot \mid x_{\le t})$ denote the model’s predictive distribution at position $t$ (obtained by softmax over the LM head logits).
For each token $t$, let $s_t \in [0,1]^F$ be the ground-truth semantic feature vector and $\hat{s}_t \in [0,1]^F$ the model’s prediction from hidden states. 
We minimize the total loss
\begin{equation}
\mathcal{L} =  
\underbrace{\mathcal{L}_{\mathrm{LM}}^{\mathrm{LS}}}_{\text{label-smoothed cross-entropy}}
  + 
\lambda_{\mathrm{aux}} \underbrace{\mathcal{L}_{\mathrm{aux}}}_{\text{BCE on }\hat{s}_t \text{ vs. } s_t}
  + 
\lambda_{\mathrm{uni}} \underbrace{\mathcal{L}_{\mathrm{uni}}}_{\text{adjective-class uniformizer}}
\end{equation}
where we use $\lambda_{\mathrm{aux}}=0.5$ and $\lambda_{\mathrm{uni}}=0.01$. BCE is the \textit{binary cross-entropy}, a negative log-likelihood for a Bernoulli target and is used when each output dimension is an independent yes/no label (e.g. each semantic feature is present or not). For a target $y\in\{0,1\}$ and a predicted probability $p\in(0,1)$, we have \footnote{For example, if a model outputs a logit $z$ with $p=\sigma(z)$, then
$\mathrm{BCE}(y,\sigma(z)) = -\big[y \log \sigma(z) + (1-y) \log\big(1-\sigma(z)\big)\big]$. This differs from (multi-class) cross-entropy, which assumes exactly one class is correct; BCE allows multiple features to be “on” simultaneously.}:
$$
\mathrm{BCE}(y,p) = -\big[y\log p + (1-y)\log(1-p)\big]
$$
For a feature vector $s_t \in [0,1]^F$ with predictions $\hat{s}_t\in(0,1)^F$, we compute BCE per feature and then average over features $f=1,\dots,F$.

\noindent\textbf{Label-smoothed LM loss.}
With label-smoothing $\varepsilon=0.02$, the smoothed target for $y_t$ over vocabulary $\mathcal{V}$ is 
$q_\varepsilon(v)= (1-\varepsilon) \mathbf{1}[v=y_t] + \varepsilon/|\mathcal{V}|$. 
The loss is the average cross-entropy over non-pad positions:
\begin{equation}
\mathcal{L}_{\mathrm{LM}}^{\mathrm{LS}}
 = 
\frac{1}{Z}\sum_{t=1}^{L}
\mathrm{CE} \big(q_\varepsilon(\cdot),  p_\theta(\cdot \mid x_{\le t})\big),
\qquad
Z=\text{\# of non-pad tokens}
\end{equation}
This smoothing reduces overconfident peaks on the target token, improving calibration and generalization, which is particularly helpful with small vocabularies and synthetic data.

\noindent\textbf{Auxiliary semantic reconstruction.}
The auxiliary head predicts $\hat{s}_t \in [0,1]^F$ and is trained with binary cross-entropy (BCE) per feature:
\begin{equation}
\mathcal{L}_{\mathrm{aux}}
 = 
\frac{1}{Z}\sum_{t=1}^{L}\sum_{f=1}^{F}
\mathrm{BCE} \big(\hat{s}_{t,f},  s_{t,f}\big)
\end{equation}
where $Z=L \cdot F$ normalizes over all token-feature pairs. 
We treat fuzzy memberships as \textit{soft} Bernoulli targets ($s_{t,f}\in[0,1]$), which standard BCE supports.
This loss term explicitly aligns hidden states with the interpretable semantic matrix $S$, reducing representation ambiguity and ensuring that the same features used for test-time control (e.g. polarity/strength) are reliably present in the model’s internal states, which in turn improves next-token prediction and stabilizes controllable decoding.

\noindent\textbf{Adjective-class uniformizer.}
Let $\mathcal{A}\subseteq\{1,\dots,L\}$ be positions whose target token is an adjective.
For $t \in \mathcal{A}$, let $\mathcal{C}_t \subseteq \mathcal{V}$ be the corresponding adjective class (positive or negative). 
Let $\ell_t \in \mathbb{R}^{|\mathcal{V}|}$ be the pre-softmax logits at $t$, and define the class-restricted distribution
$p_t^{(\mathcal{C}_t)}=\mathrm{softmax} \big(\ell_t[\mathcal{C}_t]\big)$.
We penalize deviation from the uniform distribution $u_{\mathcal{C}_t}$ on that class using the Kullback-Leibler (KL) divergence:
\begin{equation}
\mathcal{L}_{\mathrm{uni}}
 = 
\frac{1}{|\mathcal{A}|}\sum_{t\in\mathcal{A}}
\mathrm{KL} \big(p_t^{(\mathcal{C}_t)}  \big\|  u_{\mathcal{C}_t}\big)
\end{equation}
where $u_{\mathcal{C}_t}(w)=1/|\mathcal{C}_t|$ for $w \in \mathcal{C}_t$ (zero otherwise). This term acts as a mild, class-conditional entropy regularizer that prevents mode collapse onto a few frequent adjectives, thereby improving calibration and enabling controllable selection of rare or held-out adjectives at test time without materially damaging perplexity.

\paragraph{Controllable decoding with OOD-friendly sampling.}
We use a \textit{finite-state grammar} (FSG) to constrain a clause:
\[
\textsc{SUBJ}\rightarrow\textsc{VERB}\rightarrow\text{“the”}\rightarrow\textsc{OBJ}\rightarrow\text{“,”}\rightarrow\textsc{INTENS}\rightarrow\textsc{ADJ}\rightarrow\textsc{PUNCT}.
\]
At each state we apply (i) a \textit{grammar mask} that keeps only allowable tokens, (ii) a state-aware \textit{logit steer} at \textsc{ADJ}/\textsc{PUNCT} that adds small, control-driven shifts to the logits (controls are scalar features in $[0,1]$, e.g. \texttt{pos\_high} or \texttt{is\_question}), and (iii) a last-$k$ repetition penalty, with $k=3$ by default. Under strong requests we optionally \textit{hard}-restrict adjectives to the desired polarity and deterministically set punctuation (“!” or “?”). 

At the \textsc{ADJ} state, let $\mathcal{C}\subseteq\mathcal{V}$ be the indices of the active adjective class (positive or negative), and let $\ell_{\mathcal{C}}\in\mathbb{R}^{|\mathcal{C}|}$ be the class-restricted logits from the LM head. With temperature $T{>}0$ we define the pre-mix distribution $p=\mathrm{softmax}(\ell_{\mathcal{C}}/T)$. To promote OOD adjectives within the class, we sample from the convex group-uniform mixture
\begin{equation}
q  =  (1-\alpha)  p  +  \alpha  \mathrm{Unif}(\mathcal{C}), 
\qquad \alpha\in[0,1]
\end{equation}
where $\mathrm{Unif}(\mathcal{C})$ is the uniform distribution on $\mathcal{C}$:
\[
\mathrm{Unif}(\mathcal{C})(w)=
\begin{cases}
1/|\mathcal{C}|, & w\in\mathcal{C},\\
0, & \text{otherwise}.
\end{cases}
\]
We form $q$ as a convex mixture of $p$ and $\mathrm{Unif}(\mathcal{C})$ with $\alpha\in[0,1]$; in our experiments, $p$ is the class-restricted, renormalized softmax $p=\mathrm{softmax}(\ell/T)$ on indices in $\mathcal{C}$. Then we apply nucleus (top-$\rho$) sampling to $q$ with threshold $\rho\in(0,1]$. This “mix-then-truncate” step preserves probability mass for rare or held-out adjectives: the uniform component prevents collapse onto a few high-probability  \textit{seen} adjectives so held-out ones retain meaningful probability, and applying nucleus to $q$ (rather than to the raw softmax) ensures these boosted tokens survive the truncation.

\vspace{-0.25em}
\section{Tests}
\subsection{Experimental setup}

\paragraph{Data.}
We construct a synthetic corpus of clause-structured sentences. With probability $0.6$, a second clause is appended that corefers to the first-clause subject via a pronoun (“she/he/they”). The vocabulary is small and fixed: subjects \{Alice, Bob, Carol, Dave, Eve\}; verbs \{finishes, reviews, trains, starts, cooks\}; objects \{task, paper, model, project, meal\}; intensifiers \{slightly, moderately, very, extremely\}; positive-polarity adjectives (pos\footnote{We use \textit{PoS} for part-of-speech (noun/verb/adj) and \textit{pos}/\textit{neg} for positive/negative polarity. Thus “adjectives (pos)” denotes positive-polarity adjectives (not PoS tags).}) \{good, great, excellent, pleasant, wonderful\}; and negative-polarity adjectives (neg) \{bad, poor, terrible, unpleasant, awful\}. To evaluate OOD control, we hold out roughly half of the adjectives (from both polarity classes) during training while keeping them available at validation time and during controlled decoding; for example, we hold out \textit{wonderful/excellent/great} (pos) and \textit{terrible/awful/unpleasant} (neg). We generate $8,000$ training and $1,200$ validation sentences and cap the sequence length at $28$ tokens, including the BOS/EOS markers.

\paragraph{Models and training.}
Both the baseline and the fusion models use the same encoder-only Transformer with tied input/output embeddings. The backbone has hidden size $d=128$, $L=4$ encoder layers, $H=4$ attention heads per layer, feed-forward width $256$ (dropout $0.1$). We train for $6$ epochs with \textit{AdamW} \cite{Loshchilov2019decoupled} (learning rate $3\times10^{-4}$, weight decay $0.01$), batch size $64$, a warmup$+$cosine learning-rate schedule (10\% warmup), gradient clipping at norm $1.0$, and label smoothing $\varepsilon=0.02$ on the LM loss. The fusion variant adds a linear projection of the semantic features and a learned gate for fusion, plus an auxiliary head that reconstructs semantics (weight $\lambda_{\text{aux}}=0.5$). \textit{Both} models are trained with a small adjective-class \textit{uniformizer} (weight $\lambda_{\text{uni}}=0.01$) that encourages within-class coverage.

\noindent\textbf{Evaluation.} We report: overall perplexity (PPL; evaluated without label smoothing), \textit{seen-only} PPL computed after masking out held-out adjectives (see Appendix.\ref{app:seenonly_PPL}); mean-squared error (MSE) of the auxiliary semantic predictions; per-token cross-entropy on a set of focus tokens (e.g., intensifiers and punctuation); control success rates for adjective polarity and punctuation; confusion matrices (intended vs.\ realized polarity); and the OOD control \textit{hit rate}, i.e., the fraction of trials where the generator selects a held-out adjective when asked for a given polarity class.

To prevent the fusion model from gaining an advantage via decoding alone, the \textit{baseline} uses the same finite-state grammar and a last-$k$ repetition penalty \footnote{In our samples, the baseline uses a stronger last-3 penalty and top-$k=20$, while Fusion uses a moderate penalty and nucleus only. See Appendix.\ref{app:impl} for details.} during generation (Section.\ref{sec:method}).
This “fair” baseline ensures that qualitative differences are attributable to the semantic channel rather than grammar constraints.

\vspace{-0.25em}
\subsection{Results}

\begin{table}[H]
\centering
\footnotesize
\begin{tabular}{lcccccc}
\toprule
Model & PPL $\downarrow$ & Seen-only PPL $\downarrow$ & Sem.\ MSE $\downarrow$ & Adj.\ ctrl.\ acc.\ $\uparrow$ & Punct.\ ctrl.\ acc.\ $\uparrow$ & OOD hit (POS/NEG) $\uparrow$\\
\midrule
Baseline & $2.249$ & $1.511$ & -- & -- & -- & -- \\
semantic fusion & $\mathbf{2.152}$ & $\mathbf{1.431}$ & $0.0087$ & $1.00$ & $1.00$ & $0.62$ / $0.43$ \\
\bottomrule
\end{tabular}
\caption{Main metrics on the synthetic task (validation). “Seen-only” excludes held-out adjectives. OOD hit is the probability of producing a held-out adjective under strong class-appropriate control.}
\label{tab:main}
\end{table}

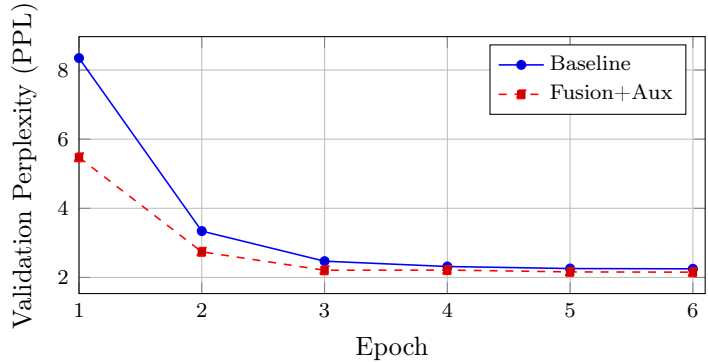
\begin{figure}[H]
\centering
\begin{tikzpicture}
\begin{axis}[
    width=0.6\linewidth,
    height=5cm,
    xlabel={Epoch},
    ylabel={Validation Perplexity (PPL)},
    xmin=1, xmax=6.1,
    grid=both,
    legend pos=north east,
    legend cell align=left,
    ylabel near ticks,
    xlabel near ticks,
    ymajorgrids=true,
    xmajorgrids=true,
    label style={font=\small},
    tick label style={font=\scriptsize},
    legend style={font=\scriptsize},
]
% Baseline
\addplot+[mark=*, semithick, mark size=1.6pt] coordinates {
    (1, 8.346) (2, 3.341) (3, 2.470) (4, 2.316) (5, 2.257) (6, 2.249)
};
\addlegendentry{Baseline}

% Fusion+Aux
\addplot+[mark=square*, dashed, semithick, mark size=1.6pt] coordinates {
    (1, 5.474) (2, 2.741) (3, 2.208) (4, 2.213) (5, 2.160) (6, 2.152)
};
\addlegendentry{Fusion+Aux}
\end{axis}
\end{tikzpicture}
\caption{Validation perplexity across epochs for Baseline and Fusion+Aux models.}
\label{fig:val-ppl-curves}
\end{figure}

\paragraph{Language modelling.}
Fusion reduces overall PPL from $2.249$ to $2.152$ ($\approx 4.3\%$ relative) and seen-only PPL from $1.511$ to $1.431$ ($\approx 5.3\%$), indicating that the semantic channel contributes predictive structure beyond what the backbone infers, even when held-out adjectives are excluded.

\paragraph{Training dynamics.}
Fig.\ref{fig:val-ppl-curves} plots validation perplexity (evaluated without label smoothing) across epochs. Both models converge smoothly, with semantic fusion below the baseline from the first epoch (5.474 vs. 8.346) and retaining a gap through epoch 6 (2.152 vs. 2.249). Most gains accrue early (by epoch 3 Fusion reaches 2.208), after which both curves plateau. A small wobble at epoch 4 for Fusion (2.208$\to$2.213) is within normal validation noise.

\paragraph{Controllability.}
Under \textit{hard} control, adjective polarity and punctuation both reach $100\%$ success (200/200 for POS and NEG; punctuation also perfect), and the intended-vs. realized confusion matrix is strictly diagonal. This demonstrates precise, user-controllable generation when explicit constraints are applied (Tables.\ref{tab:hard-control} and \ref{tab:confusion}).

\begin{table}[H]
\centering
\small
\begin{tabular}{lcc}
\toprule
Setting (N=200) & Adj.\ control acc.\ $\uparrow$ & Punct.\ control acc.\ $\uparrow$ \\
\midrule
Positive control (\texttt{pos\_high}) & $1.00$  (200/200) & $1.00$  (200/200; “!”) \\
Negative+question (\texttt{neg\_high}+\texttt{is\_question}) & $1.00$  (200/200) & $1.00$  (200/200; “?”) \\
\bottomrule
\end{tabular}
\caption{Hard-control success for the \textit{fusion} generator. Each row evaluates \(N=200\) completions under the indicated control. \textit{Adj.\ control acc.} = fraction whose final adjective belongs to the requested polarity class; \textit{Punct.\ control acc.} = fraction whose final punctuation matches the request. Hard control enforces class restriction at \textsc{ADJ} and deterministic punctuation at \textsc{PUNCT} within the one-clause FSG. Both settings achieve \(1.00\) (200/200), i.e. perfect execution.}
\label{tab:hard-control}
\end{table}

\begin{table}[H]
\centering
\small
\begin{tabular}{lccc}
\toprule
 & \multicolumn{3}{c}{\textbf{Realized}}\\
\cmidrule(lr){2-4}
\textbf{Intended} & POS & NEG & OTHER \\
\midrule
POS & 200  (1.00) & 0  (0.00) & 0  (0.00) \\
NEG & 0  (0.00) & 200  (1.00) & 0  (0.00) \\
\bottomrule
\end{tabular}
\caption{Confusion matrix under \textit{hard control} with the fusion generator. Rows are the \textit{intended} adjective polarity; columns are the \textit{realized} polarity. “OTHER” denotes any adjective not in the POS/NEG lists. Each row aggregates \(N=200\) generations. The strictly diagonal matrix (200 on-class, 0 off-class) indicates no polarity flips and no off-class/invalid adjectives.}
\label{tab:confusion}
\end{table}

\paragraph{Out-of-distribution control.}
With half the adjectives held out per polarity, the fusion model selects held-out \textit{positive} adjectives in $62\%$ of controlled runs and \textit{negative} held-outs in $43\%$ (Table.\ref{tab:ood}). The positive rate is on par with a class-uniform reference of $60\%$ (3/5 held-out), indicating class-level generalization rather than memorization. The lower NEG rate likely reflects (i) a less aggressive uniform-mixture setting and slightly stronger nucleus truncation during NEG decoding, and (ii) the use of medium strength (\texttt{str\_med}), which co-occurs more with seen negatives (\textit{bad}/\textit{poor}) than with stronger held-outs (\textit{awful}/\textit{terrible}). These behaviors are consistent with the uniformizer and group-uniform mixture in  Section.\ref{sec:method}; in practice, we can increase the NEG mix weight $\alpha$ (and/or relaxing nucleus on the mixture) and steer to higher strength to raise the NEG OOD hit rate.

\begin{table}[H]
\centering
\small
\begin{tabular}{lcc}
\toprule
Class (held-out set) & OOD hit rate $\uparrow$ & Trials \\
\midrule
POS (\texttt{wonderful, excellent, great}) & $0.62$ & 200 \\
NEG (\texttt{terrible, awful, unpleasant}) & $0.43$ & 200 \\
\bottomrule
\end{tabular}
\caption{OOD control: probability of producing a \textit{held-out} adjective under the correct class control.}
\label{tab:ood}
\end{table}

\paragraph{Qualitative generations (unprompted).}
We first show \textit{unprompted} one-clause generations that start from \texttt{<bos>} (no prefix). Some examples are shown in Table.\ref{fig:qual}. It is evident that, neutral samples are grammatical and coherent; under positive control the model reliably realizes high-intensity positive adjectives (e.g. \textit{“extremely wonderful!”}), while negative+question control yields consistent interrogatives (e.g. \textit{“moderately bad?”}). The baseline, even with the fair grammar, tends to repeat safe patterns (e.g. \textit{“Dave starts the paper, moderately good.”}), whereas the fusion model exhibits more varied adjective choices consistent with the requested semantics. 

\begin{figure}[H]
\centering
\small
\begin{tabular}{p{0.9\linewidth}}
\toprule
\textbf{Baseline (fair)}: \textit{Dave starts the paper, moderately good.}\\
\midrule
\textbf{Fusion (neutral)}:\\
\textit{• Alice cooks the model, moderately pleasant!}\\
\textit{• Alice reviews the model, moderately pleasant!}\\[0.25em]
\textbf{Fusion (positive \& strong)}:\\
\textit{• Alice reviews the paper, moderately excellent!}\\
\textit{• Carol finishes the model, extremely wonderful!}\\
\textit{• Alice reviews the model, moderately pleasant!}\\[0.25em]
\textbf{Fusion (negative \& question)}:\\
\textit{• Carol finishes the model, moderately bad?}\\
\textit{• Alice cooks the model, moderately terrible?}\\
\textit{• Alice reviews the model, extremely awful?}\\
\bottomrule
\end{tabular}
\caption{Representative \textit{unprompted} generations under different controls (decoding starts at \texttt{<bos>}). Each Fusion block shows three independent samples. All runs use the one-clause finite-state grammar and nucleus sampling ($p=0.9$). \textit{Baseline (fair)}: temperature $0.7$, top-$k=20$, last-$3$ repetition penalty (strong). \textit{Fusion (neutral)}: temperature $0.7$, no control. \textit{Fusion (positive)}: control \{\texttt{pos\_high}$=0.95$, \texttt{str\_high}$=0.9$\}, with “!” at \textsc{PUNCT}. \textit{Fusion (negative \& question)}: control \{\texttt{neg\_high}$=0.95$, \texttt{is\_question}$=1.0$, \texttt{str\_med}$=0.6$\}, with “?” at \textsc{PUNCT}. 
“Baseline (fair)” denotes decoding under the same FSG, sampling settings, and repetition guard as Fusion, but \textit{without} semantic steering or class-mixture, ensuring differences are not due to decoding.}
\label{fig:qual}
\end{figure}

\paragraph{Prompted continuation (prefix-conditioned).} 
In a separate setting we provide a fixed prefix that matches the grammar (e.g. “Carol starts the model,”). This \textit{prompted} case is different from Fig.\ref{fig:qual}: here decoding begins at \textsc{INTENS} given the prefix, so the subject/verb/object are held fixed by the prompt. The results presentedin Fig.\ref{tab:prompted} shows that, the baseline continues with safe completions (e.g. “slightly good.”), whereas the fusion model yields “extremely pleasant!” under no control and respects requested controls (e.g. “extremely excellent!” or “extremely bad?”).

\begin{table}[H]
\centering
\small
\begin{tabular}{p{0.9\linewidth}}
\toprule
\textbf{Prompt (fixed prefix)}: \textit{Carol starts the model,} \\
\midrule
\textbf{Baseline (fair)}: \textit{Carol starts the model, slightly good.}\\
\textbf{Fusion (no control)}: \textit{Carol starts the model, extremely pleasant!}\\
\textbf{Fusion (positive \& strong)}: \textit{Carol starts the model, extremely excellent!}\\
\textbf{Fusion (negative \& question)}: \textit{Carol starts the model, extremely bad?}\\
\bottomrule
\end{tabular}
\caption{Prefix-conditioned completions with a fixed prefix. Unlike Fig.\ref{fig:qual}, which shows \textit{unprompted} generations (no prefix), these examples condition on the prefix and begin decoding at \textsc{INTENS}. Settings: baseline $T=0.7$, top-$p=0.9$, top-$k=20$; fusion $T=0.7$, top-$p=0.9$; controls as indicated.}
\label{tab:prompted}
\end{table}

\paragraph{Focus-CE analysis.}
Table.\ref{tab:focus-ce} reports per-occurrence token-wise cross-entropy (CE). semantic fusion substantially improves key control and boundary tokens: \texttt{!} ($-34.8\%$), \texttt{?} ($-26.1\%$), \texttt{very} ($-30.8\%$), \texttt{good} ($-31.4\%$), and \texttt{slightly} ($-17.9\%$) relative to the baseline. For held-out adjectives, CE remains high as expected; \texttt{great} (held-out POS) increases ($+30.3\%$), reflecting a trade-off where the uniformizer flattens within-class distributions to improve \textit{coverage} for OOD control rather than concentrating probability on any single held-out item. Additionally, \texttt{terrible} (held-out NEG) shows a small improvement ($-0.5\%$). A minor regression on commas ($+21.1\%$) has negligible impact on controllability or well-formedness.

\begin{table}[H]
\centering
\small
\setlength{\tabcolsep}{8pt}
\begin{tabular}{lcccc}
\toprule
Token & Baseline CE $\downarrow$ & Fusion CE $\downarrow$ & $\Delta$ (\%) $\downarrow$ & Hold-out? \\
\midrule
\texttt{good}      & 0.00258 & 0.00177 & $-31.4$ & No \\
\texttt{great}     & 6.86868 & 8.94856 & $+30.3$ & Yes (POS) \\
\texttt{terrible}  & 7.02154 & 6.98764 & $-0.5$  & Yes (NEG) \\
\texttt{slightly}  & 0.00309 & 0.00254 & $-17.9$ & No \\
\texttt{very}      & 0.00277 & 0.00191 & $-30.8$ & No \\
\texttt{!}         & 4.42129 & 2.88172 & $-34.8$ & No \\
\texttt{?}         & 3.94640 & 2.91456 & $-26.1$ & No \\
\texttt{,}         & 0.00293 & 0.00354 & $+21.1$ & No \\
\bottomrule
\end{tabular}
\caption{Focus-token cross-entropy (CE; lower better, in \textit{nats}) on the validation set, computed \textit{per occurrence} of the token (i.e. conditioning on positions where that token is the gold next token). $\Delta$ is the relative change $(\text{Fusion}-\text{Baseline})/\text{Baseline}\times 100\%$. Held-out adjectives in this run: \textit{POS}=\{\texttt{wonderful}, \texttt{excellent}, \texttt{great}\}, \textit{NEG}=\{\texttt{terrible}, \texttt{awful}, \texttt{unpleasant}\}.}
\label{tab:focus-ce}
\end{table}

\paragraph{Semantic reconstruction.}
The auxiliary head attains an MSE of $0.0087$ (averaged over non-pad token-feature pairs), confirming that it reconstructs the fuzzy-membership vector and the hidden states encode the fuzzy semantics with low error, which in turn supports reliable test-time control.

We therefore observe that, semantic fusion (i) improves PPL and seen-only PPL with minimal overhead, (ii) delivers perfect execution under hard polarity/punctuation control, (iii) generalizes control to held-out adjectives at substantial rates, and (iv) reduces token-level CE where control matters most (intensity and punctuation), with minor, explainable regressions on certain held-out or low-salience tokens.

\vspace{-0.25em}
\section{Discussion}

Semantic fusion improves language modeling accuracy by injecting an explicit, interpretable semantic channel alongside the text stream. The sentence-level matrix \(S\) provides structure at positions with salient choices for this task (e.g. \textsc{INTENS}, \textsc{ADJ}, \textsc{PUNCT}), allowing the backbone to condition on cues it would otherwise have to infer implicitly. A lightweight gated adapter learns to weight this signal where helpful and ignore it where not, while the auxiliary head (with MSE \(=0.0087\)) encourages hidden states to align with \(S\). Empirically, this corresponds to lower validation perplexity overall and on the \textit{seen-only} subset, and to reduced token-wise cross-entropy on intensity and punctuation tokens, indicating that gains are not confined to directly supervised adjectives but reflect a broader inductive bias.

The same semantic predicates that the model learns to reconstruct at training time are reused at test time to steer decoding, which makes control robust and transparent. A finite-state \textit{grammar mask} enforces the one-clause template; \textit{state-aware} logit steering targets the adjective and punctuation states; and fuzzy triplets (e.g. \texttt{str\_low/med/high}) expose graded knobs rather than brittle one-hot tags. Under \textit{hard} constraints (class restriction and deterministic punctuation), polarity and punctuation control are perfect (see Table.\ref{tab:hard-control}; Table.\ref{tab:confusion} being strictly diagonal). Without hard restriction, \textit{soft} steering is qualitatively effective while preserving diversity (Fig.\ref{fig:qual}).

Beyond accuracy and control, the method generalizes to OOD adjectives by promoting \textit{class semantics} rather than memorized lexemes. A small adjective-class uniformizer discourages within-class mode collapse, and group-uniform mixture sampling (with nucleus applied \textit{after} mixing) preserves probability mass for rare/held-out items; together they yield OOD hit rates of \(62\%\) for positive and \(43\%\) for negative adjectives (Table.\ref{tab:ood}). The positive rate is on par with a class-uniform reference (3 of 5 held-out items), while the lower negative rate likely reflects milder mixture/truncation settings and the use of medium strength (\texttt{str\_med}). Practically, the approach is simple and compatible with standard encoder-only LMs (weight tying retained by injecting semantics pre-encoder) and common optimization (\textit{AdamW}, label smoothing, cosine schedule), and it remains interpretable: the feature bank exposes human-readable controls and the auxiliary head’s low MSE (\(0.0087\)) quantifies how well those semantics are encoded.

\vspace{-0.25em}
\section{Conclusion, limitations and future work}

\paragraph{Conclusion.}
We introduced  \textit{semantic fusion}, a lightweight mechanism that augments a Transformer LM with a parallel, fuzzy-membership feature channel fused by a gated adapter and regularized with an auxiliary reconstruction head plus a small adjective-class uniformizer. On a controlled synthetic task, it consistently lowers validation perplexity, delivers perfect polarity and punctuation control under hard constraints, generalizes control to held-out adjectives at substantial rates, and yields remarkable reductions in token-level cross-entropy for intensity and punctuation. These gains come with minimal architectural overhead, preserve tied input-output embeddings, and align training-time supervision with test-time controls, providing an interpretable pathway for conditioned natural language generation.

\paragraph{Limitations.}
Our evaluation uses a synthetic, templated corpus with a small vocabulary and a one-clause decoding grammar (for clarity), which limits ecological validity. The semantic features are hand-specified and depend on fuzzy heuristics (e.g. polarity/strength maps and fixed membership kernels) and our strongest controllability results use \textit{hard} class restriction at decoding; purely \textit{soft} steering on natural text is more challenging. We also report automatic metrics only; human evaluations of fluency, faithfulness, and perceived control are absent. Finally, while training data often contain two clauses, our qualitative decoding demonstrations predominantly use one-clause FSG, and we did not explore prefix-conditioned prompting in the main experiments.

\paragraph{Future work.}
To scale beyond synthetic data, we plan to (i) induce the semantic channel from raw text via weak supervision (e.g. lightweight taggers, distant lexicons) or an end-to-end auxiliary predictor trained jointly; (ii) expand the feature bank to tense/aspect, modality, discourse markers, factuality/hedging, coreference, and pragmatic cues; and (iii) replace fixed memberships with learnable, monotonic kernels (e.g. temperature-annealed power laws or spline-based functions) with calibration. For controllability, we aim to strengthen  \textit{soft-only} control using contrastive or energy-based heads (or classifier-free-style logit fusion) at designated states, infer grammar state from prefixes, and move beyond a hand-crafted FSG to learned multi-clause constraints. Broader evaluation will include human studies, robustness/safety analyses (e.g. bias-aware features and guardrails), systematic ablations of the gate/auxiliary/uniformizer components, and comparisons against controllable-LM baselines (e.g. CTRL \cite{keskar2019ctrl}, PPLM \cite{dathathri2020plug}, GeDi \cite{krause2021gedi}, DExperts \cite{liu2021dexperts}, etc). Finally, we will transfer the approach to open-domain corpora and larger backbones (including causal decoders), explore multilingual settings, and integrate with parameter-efficient adapters, where semantic fusion could complement prompt-based and RLHF-style conditioning.

\vspace{-0.25em}
\section{Related work}

\paragraph{Transformer LMs and training practices.}
Modern language modeling is dominated by the \textit{Transformer} architecture \cite{vaswani2017attention}, whose scalability and capacity to learn from data with minimal inductive bias have enabled strong generative performance across domains such as machine translation \cite{wang2019learningdeep}, abstractive summarization \cite{Pilault2020extractive,kumar2023abstractive}, open-domain dialogue \cite{roller2020recipes}, and code generation \cite{chen2021evaluating}, etc. Several training practices, as adopted in our work, are standard: tying the input and output embeddings reduces parameters and often improves perplexity \cite{shehper2025,press2017using,inan2017tying}; label smoothing mitigates overconfidence and improves generalization \cite{szegedy2016rethinking,pereyra2017regularizing}; \textit{AdamW} \cite{Loshchilov2019decoupled} decouples weight decay from the adaptive update and is now a default optimizer for LMs \cite{Guan2023weighte}; warmup schedules stabilize early training for Transformers \cite{vaswani2017attention}; and cosine annealing (with or without restarts) is a widely used schedule for large-scale models \cite{loshchilov2017sgdr}. In our work, we integrate our semantic channel \textit{before} the encoder and keep the hidden width unchanged, so weight tying remains valid by construction \cite{press2017using,inan2017tying}.

\paragraph{Controllable generation and decoding.}
Control over style or attributes has been pursued via control codes (conditioning on special tokens) \cite{keskar2019ctrl}, plug-and-play guidance with external discriminators that steer logits at inference \cite{dathathri2020plug}, and classifier-guided priors such as GeDi \cite{krause2021gedi}. Orthogonal to these, constrained decoding restricts the hypothesis space using lexical or grammar constraints \cite{hokamp2017lexically}. To maintain fluency and reduce degeneration, sampling strategies such as top-$k$ \cite{fan2018hierarchical} and nucleus (top-$p$) \cite{holtzman2020curious} sampling are commonly used \cite{holtzman2020curious}, along with repetition penalties and coverage-style heuristics \cite{holtzman2020curious}. Our approach differs from prior control methods by (i) exposing an \textit{interpretable} feature channel (polarity, strength, roles) that the model learns to predict and (ii) reusing those same features for test-time steering, while optionally combining with finite-state grammar masks and nucleus sampling for reliability.

\paragraph{Fuzzy logic and auxiliary supervision.}
Fuzzy sets provide graded membership functions for linguistic categories \cite{zadeh1965fuzzy}, with standard triangular/Gaussian/trapezoidal parametrizations widely used in fuzzy systems \cite{klir1995fuzzy}. We draw on this formalism to encode token-level predicates as differentiable membership degrees, supplying a low-variance supervisory signal aligned with human-interpretable semantics. Using auxiliary objectives to shape internal representations traces back to multitask learning \cite{caruana1997multitask}; in this work, our auxiliary head predicts the fuzzy feature bank, encouraging hidden states to retain interpretable cues that can be leveraged both for language modeling and for controllable decoding. Finally, our adjective-class “uniformizer” connects to confidence-penalty/entropy-style regularization that discourages overconfident, peaky distributions \cite{pereyra2017regularizing}, but is applied \textit{class-conditionally} to promote coverage within a semantic class, aiding OOD control.

\section*{Code Availability}
The code used in this work is available at: \url{https://github.com/YongchaoHuang/semantic_fusion}

\bibliographystyle{plain}
\bibliography{reference}

\begin{thebibliography}{10}

\bibitem{caruana1997multitask}
Rich Caruana.
\newblock Multitask learning.
\newblock {\em Mach. Learn.}, 28(1):41–75, July 1997.

\bibitem{chen2021evaluating}
Mark Chen, Jerry Tworek, Heewoo Jun, Qiming Yuan, Henrique~Ponde de~Oliveira~Pinto, Jared Kaplan, Harri Edwards, Yuri Burda, Nicholas Joseph, Greg Brockman, Alex Ray, Raul Puri, Gretchen Krueger, Michael Petrov, Heidy Khlaaf, Girish Sastry, Pamela Mishkin, Brooke Chan, Scott Gray, Nick Ryder, Mikhail Pavlov, Alethea Power, Lukasz Kaiser, Mohammad Bavarian, Clemens Winter, Philippe Tillet, Felipe~Petroski Such, Dave Cummings, Matthias Plappert, Fotios Chantzis, Elizabeth Barnes, Ariel Herbert-Voss, William~Hebgen Guss, Alex Nichol, Alex Paino, Nikolas Tezak, Jie Tang, Igor Babuschkin, Suchir Balaji, Shantanu Jain, William Saunders, Christopher Hesse, Andrew~N. Carr, Jan Leike, Josh Achiam, Vedant Misra, Evan Morikawa, Alec Radford, Matthew Knight, Miles Brundage, Mira Murati, Katie Mayer, Peter Welinder, Bob McGrew, Dario Amodei, Sam McCandlish, Ilya Sutskever, and Wojciech Zaremba.
\newblock Evaluating large language models trained on code, 2021.

\bibitem{dathathri2020plug}
Sumanth Dathathri, Andrea Madotto, Janice Lan, Jane Hung, Eric Frank, Piero Molino, Jason Yosinski, and Rosanne Liu.
\newblock Plug and play language models: A simple approach to controlled text generation, 2020.

\bibitem{ethayarajh2019contextual}
Kawin Ethayarajh.
\newblock How contextual are contextualized word representations? comparing the geometry of bert, elmo, and gpt-2 embeddings, 2019.

\bibitem{fan2018hierarchical}
Angela Fan, Mike Lewis, and Yann Dauphin.
\newblock Hierarchical neural story generation, 2018.

\bibitem{Guan2023weighte}
Lei Guan.
\newblock Weight prediction boosts the convergence of adamw, 2023.

\bibitem{hewitt2019structural}
John Hewitt and Christopher~D. Manning.
\newblock {A} structural probe for finding syntax in word representations.
\newblock In Jill Burstein, Christy Doran, and Thamar Solorio, editors, {\em Proceedings of the 2019 Conference of the North {A}merican Chapter of the Association for Computational Linguistics: Human Language Technologies, Volume 1 (Long and Short Papers)}, pages 4129--4138, Minneapolis, Minnesota, June 2019. Association for Computational Linguistics.

\bibitem{hokamp2017lexically}
Chris Hokamp and Qun Liu.
\newblock Lexically constrained decoding for sequence generation using grid beam search, 2017.

\bibitem{holtzman2020curious}
Ari Holtzman, Jan Buys, Li~Du, Maxwell Forbes, and Yejin Choi.
\newblock The curious case of neural text degeneration, 2020.

\bibitem{inan2017tying}
Hakan Inan, Khashayar Khosravi, and Richard Socher.
\newblock Tying word vectors and word classifiers: A loss framework for language modeling, 2017.

\bibitem{keskar2019ctrl}
Nitish~Shirish Keskar, Bryan McCann, Lav~R. Varshney, Caiming Xiong, and Richard Socher.
\newblock Ctrl: A conditional transformer language model for controllable generation, 2019.

\bibitem{klir1995fuzzy}
George~J. Klir and Bo~Yuan.
\newblock {\em Fuzzy sets and fuzzy logic: theory and applications}.
\newblock Prentice-Hall, Inc., USA, 1994.

\bibitem{krause2021gedi}
Ben Krause, Akhilesh~Deepak Gotmare, Bryan McCann, Nitish~Shirish Keskar, Shafiq Joty, Richard Socher, and Nazneen~Fatema Rajani.
\newblock Gedi: Generative discriminator guided sequence generation.
\newblock In {\em Findings of Empirical Methods in Natural Language Processing (Findings of EMNLP)}, 2021.

\bibitem{kumar2023abstractive}
Sandeep Kumar and Anil Solanki.
\newblock An abstractive text summarization technique using transformer model with self-attention mechanism.
\newblock {\em Neural Computing and Applications}, 35:18603--18622, 2023.
\newblock Published: June 1, 2023; Issue Date: September 2023.

\bibitem{liu2021dexperts}
Alisa Liu, Maarten Sap, Ximing Lu, Swabha Swayamdipta, Chandra Bhagavatula, Noah~A. Smith, and Yejin Choi.
\newblock Dexperts: Decoding-time controlled text generation with experts and anti-experts, 2021.

\bibitem{loshchilov2017sgdr}
Ilya Loshchilov and Frank Hutter.
\newblock Sgdr: Stochastic gradient descent with warm restarts, 2017.

\bibitem{Loshchilov2019decoupled}
Ilya Loshchilov and Frank Hutter.
\newblock Decoupled weight decay regularization.
\newblock In {\em International Conference on Learning Representations (ICLR)}, 2019.
\newblock arXiv:1711.05101.

\bibitem{mikolov2013efficient}
Tomas Mikolov, Kai Chen, Greg Corrado, and Jeffrey Dean.
\newblock Efficient estimation of word representations in vector space, 2013.

\bibitem{pereyra2017regularizing}
Gabriel Pereyra, George Tucker, Jan Chorowski, Łukasz Kaiser, and Geoffrey Hinton.
\newblock Regularizing neural networks by penalizing confident output distributions, 2017.

\bibitem{peters2018deep}
Matthew~E. Peters, Mark Neumann, Mohit Iyyer, Matt Gardner, Christopher Clark, Kenton Lee, and Luke Zettlemoyer.
\newblock Deep contextualized word representations, 2018.

\bibitem{Pilault2020extractive}
Jonathan Pilault, Raymond Li, Sandeep Subramanian, and Chris Pal.
\newblock On extractive and abstractive neural document summarization with transformer language models.
\newblock In Bonnie Webber, Trevor Cohn, Yulan He, and Yang Liu, editors, {\em Proceedings of the 2020 Conference on Empirical Methods in Natural Language Processing (EMNLP)}, pages 9308--9319, Online, November 2020. Association for Computational Linguistics.

\bibitem{press2017using}
Ofir Press and Lior Wolf.
\newblock Using the output embedding to improve language models, 2017.

\bibitem{roller2020recipes}
Stephen Roller, Emily Dinan, Naman Goyal, Da~Ju, Mary Williamson, Yinhan Liu, Jing Xu, Myle Ott, Kurt Shuster, Eric~M. Smith, Y-Lan Boureau, and Jason Weston.
\newblock Recipes for building an open-domain chatbot, 2020.

\bibitem{shehper2025}
Ali Shehper, Anibal~M. Medina-Mardones, Lucas Fagan, Bartłomiej Lewandowski, Angus Gruen, Yang Qiu, Piotr Kucharski, Zhenghan Wang, and Sergei Gukov.
\newblock What makes math problems hard for reinforcement learning: a case study, 2025.

\bibitem{su2023roformer}
Jianlin Su, Yu~Lu, Shengfeng Pan, Ahmed Murtadha, Bo~Wen, and Yunfeng Liu.
\newblock Roformer: Enhanced transformer with rotary position embedding, 2023.

\bibitem{szegedy2016rethinking}
Christian Szegedy, Vincent Vanhoucke, Sergey Ioffe, Jonathon Shlens, and Zbigniew Wojna.
\newblock Rethinking the inception architecture for computer vision, 2015.

\bibitem{tenney2019bert}
Ian Tenney, Dipanjan Das, and Ellie Pavlick.
\newblock Bert rediscovers the classical nlp pipeline, 2019.

\bibitem{vaswani2017attention}
Ashish Vaswani, Noam Shazeer, Niki Parmar, Jakob Uszkoreit, Llion Jones, Aidan~N. Gomez, Lukasz Kaiser, and Illia Polosukhin.
\newblock Attention is all you need, 2023.

\bibitem{wang2019learningdeep}
Qiang Wang, Bei Li, Tong Xiao, Jingbo Zhu, Changliang Li, Derek~F. Wong, and Lidia~S. Chao.
\newblock Learning deep transformer models for machine translation.
\newblock In Anna Korhonen, David Traum, and Llu{\'i}s M{\`a}rquez, editors, {\em Proceedings of the 57th Annual Meeting of the Association for Computational Linguistics}, pages 1810--1822, Florence, Italy, July 2019. Association for Computational Linguistics.

\bibitem{zadeh1965fuzzy}
L.A. Zadeh.
\newblock Fuzzy sets.
\newblock {\em Information and Control}, 8(3):338--353, 1965.

\end{thebibliography}

\newpage
\appendix

\section{Implementation details}
\label{app:impl}

\paragraph{Compute environment.}
Experiments were run in Google Colab, on a Linux x86\_64 virtual machine (KVM) with a single-socket Intel Xeon CPU @ 2.20 GHz (6 physical cores / 12 threads), 179 GB RAM, and \(\sim\)253 GB local storage. The host supports AVX2/AVX-512 (incl.\ VNNI). No discrete GPU was available, so all training/inference used the PyTorch CPU backend.

\paragraph{Backbone and parameterization.}
Both baseline and fusion models use an encoder-only Transformer with \textit{tied} input/output embeddings \footnote{\textbf{Tied input/output embeddings.} The output softmax weights reuse the input embedding matrix: if $E\in\mathbb{R}^{V\times d}$ is the token embedding table, set $W_{\text{out}}=E^\top$ so logits are $z_t=h_t E^\top+b$. This reduces parameters and aligns input/output geometry; it requires hidden size $d$ to equal the embedding dimension.}.
Hidden size $d=128$, layers $L=4$, heads $H=4$, feed-forward width $256$, dropout $0.1$, sinusoidal positional encodings.

\paragraph{Semantic feature bank.}
For each token we compute $s_t \in [0,1]^F$ over interpretable predicates:
\texttt{is\_noun}, \texttt{is\_verb}, \texttt{is\_adj},
\texttt{is\_subject}, \texttt{is\_object}, \texttt{is\_head},
\texttt{is\_bos}, \texttt{is\_eos}, \texttt{is\_comma}, \texttt{is\_question},
\texttt{pos\_low/med/high}, \texttt{neg\_low/med/high},
\texttt{str\_low/med/high}, \texttt{coref\_subject},
\texttt{is\_capitalized}, \texttt{is\_pronoun}.
Graded predicates use fuzzy/triangular memberships with a power kernel
\[
\mu(x;c,\tau)=0.9^{|x-c|/\tau},\quad
\text{Tri}(x;\{0.2,0.6,1.0\},\tau),\ \ \tau=0.35
\]

\paragraph{Fusion module.}
Let $e_t=E[x_t]$, $u_t=W_s s_t$, and $g_t=\sigma \big(W_g[e_t;s_t]\big)$.
The fused input to the encoder is
\[
h_t^{(0)} = e_t + u_t + g_t\odot u_t
\]
followed by positional encoding and the Transformer encoder.
The LM head is tied: $p(y_t|x_{\le t})=\mathrm{softmax}(E^\top H_t)$.
An auxiliary MLP predicts $\hat{s}_t=\sigma(\mathrm{MLP}(H_t))$.

\paragraph{Optimization and schedule.}
We train for $6$ epochs with \textit{AdamW} \cite{Loshchilov2019decoupled} (learning rate $3\times10^{-4}$, weight decay $0.01$), batch size $64$, gradient clipping at norm $1.0$, and a warmup$+$cosine schedule (10\% warmup). Label smoothing $\varepsilon=0.02$ is applied to the LM loss during training (turned  \textit{off} for validation metrics). The auxiliary semantic head uses BCE with weight $\lambda_{\text{aux}}=0.5$. The adjective-class uniformizer has coefficient $\lambda_{\text{uni}}=0.01$:
\[
\mathcal{L}_{\mathrm{uni}}=\frac{1}{|\mathcal{A}|}\sum_{t\in\mathcal{A}}
\mathrm{KL}\!\left(p_t^{(\mathcal{C}_t)}  \middle\|  u_{\mathcal{C}_t}\right),
\qquad
u_{\mathcal{C}_t}(w)=\tfrac{1}{|\mathcal{C}_t|}
\]

\paragraph{Data generation.}
Synthetic corpus with probability $0.6$ of adding a second clause that corefers via pronoun (\texttt{she/he/they}).
Vocabulary:
subjects \{\texttt{Alice}, \texttt{Bob}, \texttt{Carol}, \texttt{Dave}, \texttt{Eve}\};
verbs \{\texttt{finishes}, \texttt{reviews}, \texttt{trains}, \texttt{starts}, \texttt{cooks}\};
objects \{\texttt{task}, \texttt{paper}, \texttt{model}, \texttt{project}, \texttt{meal}\};
intensifiers \{\texttt{slightly}, \texttt{moderately}, \texttt{very}, \texttt{extremely}\};
adjectives (pos) \{\texttt{good}, \texttt{great}, \texttt{excellent}, \texttt{pleasant}, \texttt{wonderful}\} and (neg) \{\texttt{bad}, \texttt{poor}, \texttt{terrible}, \texttt{unpleasant}, \texttt{awful}\}.
We hold out roughly half of the adjectives (both polarities) for OOD control; e.g. pos hold-out \{\texttt{wonderful}, \texttt{excellent}, \texttt{great}\}, neg hold-out \{\texttt{terrible}, \texttt{awful}, \texttt{unpleasant}\}.
Train/val sizes: $8,000$/$1,200$; max length $28$ (incl.\ BOS/EOS).

\paragraph{Prompted decoding.}
If a user-supplied prefix matches the one-clause FSG (e.g.  \textit{Carol starts the model,}), we validate it, set the next grammar state accordingly (here, \textsc{INTENS}), and continue decoding from that state. For fusion, semantic features are recomputed for the prefix before continuation.

\paragraph{Decoding.}
Finite-state grammar (FSG) for one clause:
\[
\textsc{SUBJ}\to\textsc{VERB}\to\text{“the”}\to\textsc{OBJ}\to\text{“,”}\to\textsc{INTENS}\to\textsc{ADJ}\to\textsc{PUNCT}.
\]
We apply a grammar mask (allowed tokens only), a last-$k$ repetition penalty ($k=3$; stronger for the baseline, factor $\approx 2.5$, vs.\ fusion $\approx 1.5$), and state-aware logit steering at \textsc{INTENS}/\textsc{ADJ}/\textsc{PUNCT}.
Under strong requests we optionally hard-restrict adjectives to the desired polarity and deterministically set punctuation.
To encourage OOD coverage at \textsc{ADJ} we sample from a group-uniform mixture and then apply nucleus filtering:
\[
p=\operatorname{softmax}(\ell_{\mathcal C}/T),\qquad
u_{\mathcal C}(w)=\tfrac{1}{|\mathcal C|}\mathbf 1[w\in\mathcal C],\qquad
q=(1-\alpha)\,p+\alpha\,u_{\mathcal C},\ \ \alpha\in[0,1].
\]
\[
\text{Sample using nucleus (top-}\rho\text{) applied to }q
\]
Typical settings:
baseline fair decoding uses temperature $0.7$, top-$p=0.9$, top-$k=20$;
fusion neutral uses $T=0.7$, $p=0.9$;
fusion (positive \& strong) uses control \{\texttt{pos\_high}$=0.95$, \texttt{str\_high}$=0.9$\} with mixture $T=1.5$, $p=1.0$, $\alpha=0.97$;
fusion (negative \& question) uses \{\texttt{neg\_high}$=0.95$, \texttt{is\_question}$=1.0$, \texttt{str\_med}$=0.6$\} with $T=1.3$, $p=0.95$, $\alpha=0.85$ (and at \textsc{PUNCT} choose “?” deterministically).

\paragraph{Metrics.}
Validation PPL is computed with label smoothing \textit{off}.
Seen-only PPL masks positions whose gold token is in the held-out adjective set before averaging cross-entropy (Appendix.\ref{app:seenonly_PPL}).
Semantic MSE averages $(\hat{s}_t-s_t)^2$ over non-pad positions.
Focus-token CE is the mean cross-entropy conditioned on positions where a given token is the gold target.
Control accuracy is measured over $N=200$ generations per setting (adjective polarity and punctuation).
OOD hit rate is the fraction of controlled generations that realize a \textit{held-out} adjective.

\paragraph{Seen-only PPL computation.}
Let $\mathcal{H}$ be the set of held-out adjective ids.
For each position $t$, we drop the loss if the gold token $y_t\in\mathcal{H}$ and average cross-entropy over the remaining positions; perplexity is $\exp$ of this average. See also Appendix.\ref{app:seenonly_PPL}.

\paragraph{Reproducibility \& configuration.}
We fix the random seed for \texttt{random}, \texttt{numpy}, and \texttt{torch}, and set the device to \texttt{cuda} if available, else \texttt{cpu}. All other architectural choices (tied embeddings; 4-layer encoder, $d=128$, $H=4$, FFN width $256$) and decoding settings are specified in the corresponding sections above to avoid duplication.

\section{Seen-only perplexity}
\label{app:seenonly_PPL}

This appendix formalizes the \textit{seen-only} perplexity reported in our experiments and matches the implementation of \texttt{eval\_perplexity\_seen\_only} in the code.

\paragraph{Setup and notation.}
Let $x_{1:L}$ be a validation sequence with gold next-token targets $y_{1:L}$ over vocabulary $\mathcal{V}$, and let $p_\theta(\cdot\mid x_{\le t})$ denote the model's next-token distribution (evaluated \textit{without} label smoothing). Let $\mathcal{H}\subset\mathcal{V}$ be the set of \textit{held-out} adjective token ids used for OOD testing:
\[
\mathcal{H}  =  \big\{\mathrm{stoi}(w) :  w \in \texttt{ADJ\_POS\_HOLD}\cup\texttt{ADJ\_NEG\_HOLD}\big\}
\]
Define the per-position cross-entropy (natural log)
\[
\mathrm{CE}_t  =  -\log p_\theta \big(y_t  \big|  x_{\le t}\big)
\]
and a binary weight that drops padding and held-out targets:
\[
w_t  =  \mathbf{1} \big[y_t \neq \texttt{\textless pad\textgreater}\big]\cdot \mathbf{1} \big[y_t \notin \mathcal{H}\big]
\]

\paragraph{Metric.}
Aggregate the masked loss and token count
\[
L_{\text{seen}}  =  \sum_{t=1}^{L} w_t  \mathrm{CE}_t,
\qquad
N_{\text{seen}}  =  \sum_{t=1}^{L} w_t
\]
and report
\[
\mathrm{PPL}_{\text{seen-only}}
 = 
\exp \Big(\tfrac{L_{\text{seen}}}{N_{\text{seen}}}\Big)
\]

\paragraph{Batch form.}
For a batch with logits $\mathbf{Z}\in\mathbb{R}^{B\times L\times |\mathcal{V}|}$, gold $y\in\mathcal{V}^{B\times L}$, and non-pad mask $m\in\{0,1\}^{B\times L}$, define
\[
\kappa_{b,t}=\mathbf{1}[y_{b,t}\notin \mathcal{H}],\quad
c_{b,t}=\mathrm{CE} \big(\mathrm{softmax}(\mathbf{Z}_{b,t,:}),\ y_{b,t}\big)
\]
Then
\[
\bar c_{\text{seen}}=\frac{\sum_{b,t} c_{b,t}  m_{b,t}  \kappa_{b,t}}
{\sum_{b,t} m_{b,t}  \kappa_{b,t}},
\qquad
\mathrm{PPL}_{\text{seen-only}}=\exp(\bar c_{\text{seen}})
\]

\paragraph{Implementation notes.}
(i) Label smoothing is \textit{disabled} for this evaluation (true one-hot targets).
(ii) If $\sum_{b,t} m_{b,t}\kappa_{b,t}=0$ (no eligible positions), the implementation returns \texttt{NaN}.
(iii) Context filtering is \textit{not} applied: held-out adjectives may appear in the \textit{inputs} $x_{\le t}$; we only exclude positions whose \textit{targets} are held-out.

\paragraph{Rationale.}
This metric isolates predictive quality on the \textit{seen} portion of the vocabulary under realistic contexts, which is why $\mathrm{PPL}_{\text{seen-only}}$ is typically lower than overall PPL in our runs.

\end{document}